# Collaborative model based design of automated and robotic agricultural vehicles in the Crescendo Tool

[1,3,*]Martin Peter Christiansen, [4]Morten Stiggaard Laursen, [1]Rasmus Nyholm Jørgensen, [2]Ibrahim A. Hameed

[1]Aarhus University, Department of Engineering, Finlandsgade 22, 8200 Aarhus N, Denmark
[2]Aalborg University, Department of Electronic Systems, Fredrik Bajers Vej 7B, 9220 Aalborg
[3]Conpleks Innovation, Fælledvej 17, 7600 Struer, Denmark
[4]University of Southern Denmark, Faculty of Engineering, Niels Bohrs Allé 1, 5230 Odense M, Denmark

**Abstract:** This paper describes a collaborative modelling approach to automated and robotic agricultural vehicle design. The Cresendo technology allows engineers from different disciplines to collaborate and produce system models. The combined models are called co-models and their execution co-simulation. To support future development efforts a template library of different vehicle and controllers types are provided. This paper describes a methodology to developing co-models from initial problem definition to deployment of the actual system. We illustrate the development methodology with an example development case from the agricultural domain. The case relates to an encountered speed controller problem on a differential driven vehicle, where we iterate through different candidate solutions and end up with an adaptive controller solution based on a combination of classical control and learning feedforward. The second case is an example of combining human control interface and co-simulation of agricultural robotic operation to illustrate collaborative development

## Introduction

Modelling and simulation have become an integrated part of development in many engineering disciplines [1]. A model provides the developer with a tool to experiment with system design parameters and configurations. In this context, a system is a group of interacting or independent components forming a coherent whole. Domain specific modelling software tends to focus on a subset of the engineering disciplines.
Modelling and simulation across multi disciplines and domains represent a design challenge in a single tool. Collaborative modelling (co-modelling) allows system components to be developed in different development tools and run simultaneously using co-operative simulation (co-simulation) [2]. The concepts in collaborative modelling has the capability be a supplement to the features in robotics development tools like Microsoft Developer Robotic Studio and Robot Operating System.
The CRESCENDO co-simulation technology provides a model-based approach to the engineering of embedded and robotic control systems [3]. CRESCENDO models are built in order to support various forms of analysis including static analysis and simulation - the latter is the focus here. The technology supports models where the controller and plant or environment are modelled using different specialised tools. The co-simulation is intended to allow for multidisciplinary modelling with input from domain experts on the systems components.
Co-simulation performed in other tools have also been documented in litterateur. The MODELISAR project developed the Functional Mock-up Interface (FMI) to allow for co-simulation [4]. Modelica is an object-oriented, equation-based multi-domain language for simulating controlled physical systems. A fruit picking manipulator robot have been constructed and developed based on Modelica multi-domain modelling [5].
Model development is normally constrained by resources such as time and money. The developer's main goal is to achieve a system model viable for controller development. One should remember that modelling is not an attempt to put the full reality into a model, but rather an attempt to focus on the parts relevant to the developers current case [6].  In this paper we provide co-modelling development guidelines and illustrates them using two cases from the Grassbots projects.

## Material
### The CRESCENDO tool
The CRESCENDO tool uses a combination of discrete-event (DE) modelling of a digital controller and continuous-time (CT) modelling of the plant/environment for co-simulation. The overture tool and VDM formalism combined with Mathworks Matlab models the DE controllers, and the 20-sim tool models CT components. 20-sim is a modelling and simulation tool, able to model complex multi-domain dynamic systems, such as combined mechanical, electrical and hydraulic systems. VDM Real Time (VDM-RT) [7] is the dialect used in CRESCENDO with the capabilities to describe real-time, asynchronous, object-oriented features. Matlab, VDM and 20-sim are well-established formalisms with stable tool support and a record of industry use.

### Agricultural robot co-model templates
To support new development efforts using co-simulation, we provided a number of different CT and DE models related to the agricultural domain. We provide CT models of vehicles types with differential drive and front and back wheel steering.

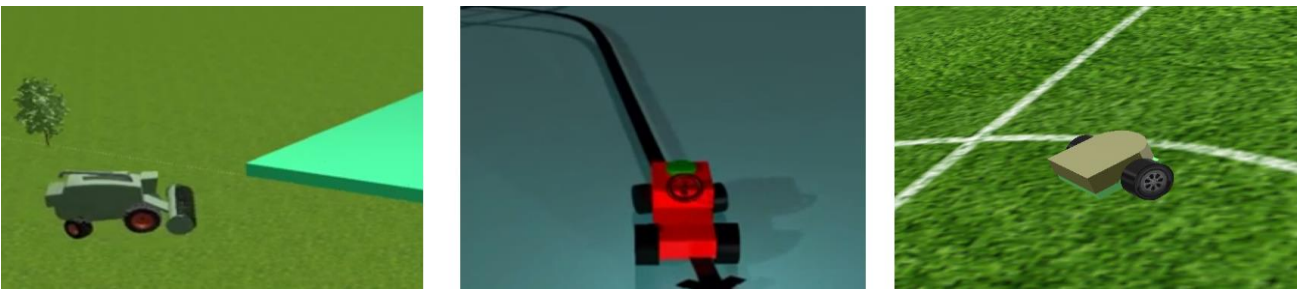

*Figure 1: Examples of visualisation the different vehicle models in the CRESENDO tool. Visualisation is not the actual model or simulation, but rather a way to present system response for a specific scenario.*

The DE model components provided in VDM is used to setup controllers for automatic steering and task operation. Dependent on the developer demands, different models is used as a basis for analysing and solving concrete problems.

## Methods
### Modelling Methodology
We propose the co-model development guidelines and process workflow in Figure 2 to encompass the full process from problem definition to deployment of a useable system. The proposed methodology used for co-model development is partly inspired by the work in [8], [9]. Where these methods mainly focus on the movement from controller requirements and environment assumptions to complete system co-model, we extend to controller deployment on the actual system.
The main focus of the co-modelling guidelines is the developers' concrete problem objective for an actual system. The development guidelines intend to gradually produce the intended system with a finer level of granularity in each stage. The development process divides the workflow into 3 main stages the co-model development are iterating trough; System Boundary Definition; Model Design; Model Deployment.
We do not claim that the output of the development process at each stage in our approach is necessarily a refinement of their predecessors. If the project stakeholders discover the need to redefine a model or problem definition, backwards movement in the development process revisiting an earlier stage occur.

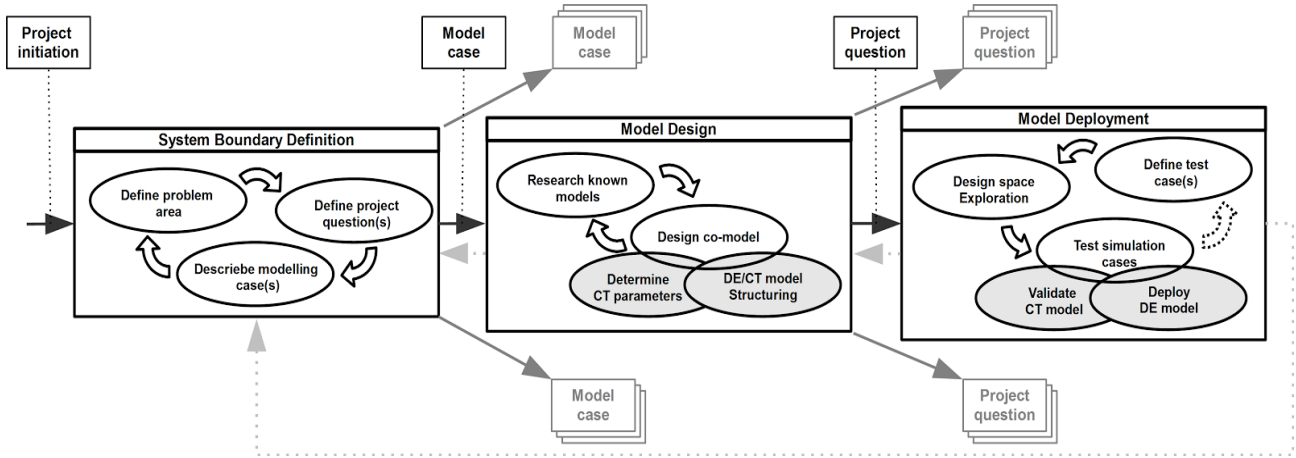

*Figure 2: The proposed co-modelling process workflow, from initial problem definitions to model deployment. Dependent on the project each stage can result in one or more branches of the next stage.*

## Results

In the Grassbots project, the grass-collector was mounted with a caterpillar track setup; controlled using feedback from rotary encoders. The encoder reports changes in position but cannot improve on the basic accuracy of the measured motion. Non-linear effects in the machine design introduces errors in the measured motion and impacting control response. When presented with problem the solution was to utilize a running average filter, to filter the effects seen in the encoder data. A first improvement was a Butterworth lowpass-filter to increase the system response capabilities. The new Butterworth filter solution was successfully deployed to the actual system. An alternative is to design a DE controller that learns the pattern of the disturbance and compensates based on learn encoder response. We illustrate her a version that learns and compensates using non-linear characteristics using an adaptive B-spline network as a feed-forward network.

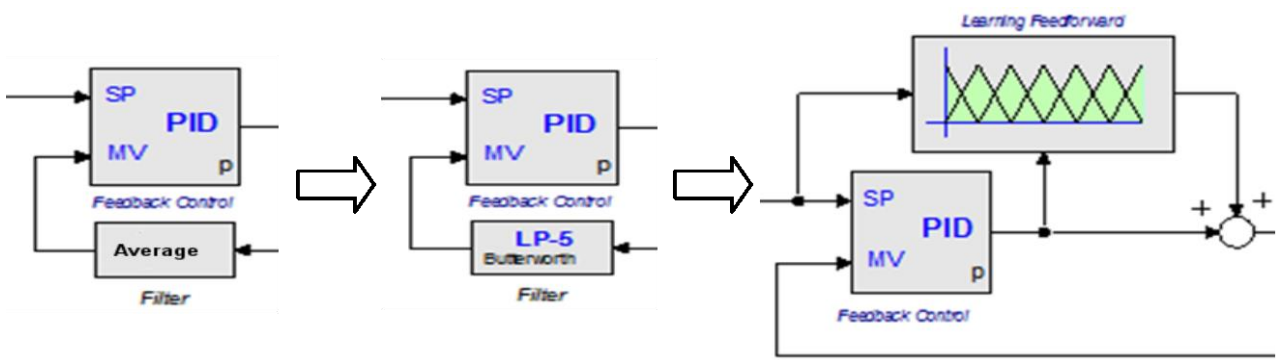

*Figure 3: The different iterations of the solution to the encoder feedback problem*

The second co-model tests inter-active route planning, where the user is able to select a certain area he/she would like to cover. Using co-simulation one can experiment with interactive software solutions before deploying them on the actual system. In the case the interactive route-planner is developed in Matlab, combined with a vehicle controller made in VDM and a vehicle model in 20-sim.
The approach to software development is also known as software in the loop testing, where the developer can verify the behaviour of production-intent code on his own computer.

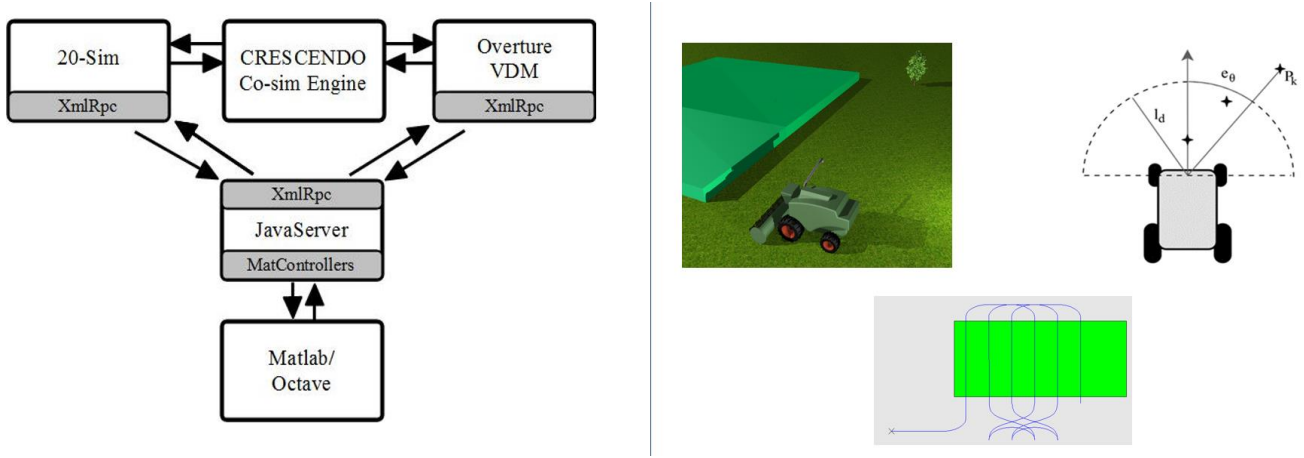

*Figure 4: The interactive route planner made using co-modelling combining components from VDM, Matlab and 20.sim.*

## Discussion

We illustrated how the guidelines can be used to develop and deploy co-models on the actual system. The case presented here should be seen as an example of what co-modelling and co-simulation can be used for. Future version of the Crescendo tool might provide modelling together with more development tools closely related to agricultural domain.

## Concluding remarks

In this paper, we have presented a methodology to develop collaborative models from the state of problem definition to actual deployment on the platform. The collaborative modelling methodology was illustrated with the two case from the GrassBots project.

## Acknowledgement

Financial support given by the Danish Ministry of Food, Agriculture and Fisheries is gratefully acknowledged. We would like to thank the Grassbots project under the EU FP7 program for partial funding and providing the example problem case. We acknowledge partial support from the EU FP7 DESTECS project on co-simulation. Thanks are due to Morten Larsen for providing measurement data for the encoder case.